\documentclass{article}
\usepackage{PRIMEarxiv}

\usepackage[utf8]{inputenc} 
\usepackage[T1]{fontenc}    
\usepackage{hyperref}       
\usepackage{url}            
\usepackage{booktabs}       
\usepackage{amsfonts}       
\usepackage{nicefrac}       
\usepackage{microtype}      
\usepackage{lipsum}
\usepackage{fancyhdr}       
\usepackage{graphicx}       
\graphicspath{{media/}}     

\usepackage{microtype}
\usepackage{graphicx}
\usepackage{subfigure}
\usepackage{booktabs} 
\newcommand{\indep}{\perp \!\!\! \perp}

\usepackage{hyperref}
\usepackage{adjustbox}
\usepackage{algorithm}
\usepackage{algpseudocode}
\usepackage{balance}
\usepackage{amsmath}
\usepackage{amssymb}
\usepackage{mathtools}
\usepackage{amsthm}
\usepackage{multirow}
\usepackage{lipsum}

\usepackage{cite}
\usepackage{hyperref}

\usepackage{footnote}
\usepackage{etoolbox}
\makeatletter
\patchcmd{\@makecaption}
  {\scshape}
  {}
  {}
  {}
\makeatother
\usepackage{xcolor}

\usepackage{amsfonts}       
\usepackage{nicefrac}       

\newtheorem{assumption}{Assumption}
\newtheorem{lemma}{Lemma}

\usepackage{bm}
\usepackage{amsmath}

\DeclareMathOperator*{\argmin}{arg\,min}
\newtheorem{theorem}{Theorem}

\usepackage{graphicx,float} 
\usepackage{amssymb}
\usepackage{algorithm}
\usepackage{amsmath,bm}
\usepackage{makecell}

\usepackage{hyperref}
\usepackage{booktabs}
\usepackage{lipsum}
\usepackage{multicol}
\usepackage[capitalize,noabbrev]{cleveref}

\theoremstyle{plain}
\theoremstyle{definition}
\theoremstyle{remark}

\usepackage{microtype}
\usepackage{graphicx}
\usepackage{subfigure}
\usepackage{booktabs} 
\usepackage{hyperref}
\usepackage{adjustbox}
\usepackage{algorithm}
\usepackage{algpseudocode}
\usepackage{balance}
\usepackage{amsmath}
\usepackage{amssymb}
\usepackage{mathtools}
\usepackage{amsthm}
\usepackage{multirow}
\usepackage{lipsum}

\usepackage{hyperref}

\usepackage{footnote}
\usepackage{etoolbox}
\makeatletter
\patchcmd{\@makecaption}
  {\scshape}
  {}
  {}
  {}
\makeatother
\usepackage{amsfonts}       
\usepackage{nicefrac}       
\usepackage{cite}
\usepackage{bm}
\usepackage{amsmath}
\usepackage{graphicx,float} 
\usepackage{amssymb}
\usepackage{algorithm}
\usepackage{amsmath,bm}
\usepackage{makecell}
\usepackage{hyperref}
\usepackage{booktabs}
\usepackage{lipsum}
\usepackage{multicol}
\usepackage[capitalize,noabbrev]{cleveref}
\usepackage[font=small]{caption} 
\DeclareMathOperator{\Var}{\textit{Var}}
\theoremstyle{plain}
\theoremstyle{definition}
\theoremstyle{remark}

\newif\ifcomments
\commentstrue
\commentsfalse
\ifcomments
    \usepackage{ulem}
    \def\picomment#1{{$\!$\color{green} [PI: #1]}}

    \def\saedit#1{{$\!$\color{blue} [SA: #1]}}

    \def\tncomment#1{{$\!$\color{red} [TN: #1]}}
    \def\tnedit#1{{$\!$\color{red} [TN: #1]}}

\else 
    \def\picomment#1{}

    \def\saedit#1{}

    \def\tncomment#1{}    
    \def\tnedit#1{}
    
\fi
\onecolumn
\begin{document}

\title{Conditional entropy minimization principle\\
           for learning domain invariant representation features}

\author{
  Thuan Nguyen \\
 Department of Computer Science \\ Tufts University \\ Medford, MA 02155 \\
  \texttt{\{Thuan.Nguyen@tufts.edu} \\
   \And
  Boyang Lyu \\
  Department of Electrical and Computer Engineering \\ Tufts University \\ Medford, MA 02155\\
  \texttt{Boyang.Lyu@tufts.edu} \\
     \And
  Prakash Ishwar \\
  Department of Electrical and Computer Engineering \\ Boston University \\ Boston, MA 02215\\
  \texttt{pi@bu.edu} \\
  \And
    Matthias Scheutz \\
 Department of Computer Science \\ Tufts University \\ Medford, MA 02155 \\
  \texttt{\{Matthias.Scheutz@tufts.edu} \\
     \And
  Shuchin Aeron \\
  Department of Electrical and Computer Engineering \\ Tufts University \\ Medford, MA 02155\\
  \texttt{Shuchin@ece.tufts.edu}
}

\maketitle

\begin{abstract}
Invariance-principle-based methods such as Invariant Risk Minimization (IRM), have recently emerged as promising approaches for Domain Generalization (DG). Despite promising theory, such approaches fail in common classification tasks due to mixing of  \textit{true invariant features} and \textit{spurious invariant features}\footnote{We use the terms ``spurious invariant features" or just ``spurious features" to denote features that are invariant across all seen domains, but change in the unseen domain \cite{chen2021iterative} \cite{rosenfeld2021risks}.}. To address this, we propose a framework based on the conditional entropy minimization (CEM) principle to filter-out the spurious invariant features leading to a new algorithm with a better generalization capability. 
We show that our proposed approach is closely related to the well-known Information Bottleneck (IB) framework and prove that under certain assumptions, entropy minimization can exactly recover the true invariant features. Our approach provides competitive classification accuracy compared to recent theoretically-principled state-of-the-art alternatives across several DG datasets\footnote{This paper was accepted at 26th International Conference on Pattern Recognition (ICPR-2022).}. 
\end{abstract}

\section{Introduction}

A fundamental assumption in most statistical machine learning algorithms is that the training data and the test data are independently and identically distributed (i.i.d). However, it  is usually violated in practice due to a phenomenon  often referred to as  the domain distribution shift where the training domain and the test domain distributions  are not the same.   This leads to an increased risk/error of the trained classifier on the test domain.   Mitigating this issue is the subject of the area broadly referred to as Domain Generalization (DG). 

Over the past decade, many methods have been proposed for DG,  under different settings \cite{wang2022generalizing} \cite{zhou2021domain}. Among these,  Invariant Risk Minimization (IRM) \cite{arjovsky2019invariant} \cite{lu2021nonlinear} has emerged as one of the promising methods. IRM is constructed based on a widely accepted assumption that the representations are general and transferable if the feature representations remain invariant from domain to domain.  However, this approach is shown to  fail in  some  simple settings where  spurious invariant features exist \cite{rosenfeld2021risks} \cite{aubin2021linear} \cite{kamath2021does} \cite{gulrajani2020search}. A particular example is the problem of classifying cow and camel images \cite{nagarajan2020understanding} \cite{bui2021exploiting} where the label is a deterministic function of the invariant features, for example, the shape of animals, and does not depend on the spurious features  such as  the background. However, because cows usually appear in a picture with a greenfield while the camels live in a desert with a yellow background, the background color could be incorrectly learned as a spurious invariant feature. This can lead to classification errors,  for example, if the cow is placed in a yellow field, then it may be misclassified as a camel. Therefore, even though an invariance-principle-based approach can learn invariant features, it may still fail in a classification task if the extracted features contain not only the true invariant features but also spurious invariant features. These spurious features could be eliminated if one can observe  a sufficiently large  number of domains \cite{chen2021iterative} \cite{rosenfeld2021risks}. For example, if the seen domain contains a picture of a cow walking in a desert.
However collecting labeled data from  all possible domains is impractical. 

Several frameworks have been proposed  to deal with the presence of spurious invariant features. For example, in \cite{ahuja2021invariance} the entropy of  the extracted features  is minimized  to filter out spurious features. However, only linear classifiers are considered and although the approach is motivated by the Information Bottleneck (IB) framework \cite{tishby2000information}, IB is not directly utilized in the learning objective.   A similar approach directly based on the IB objective function for eliminating spurious invariant features  appears  in  \cite{li2021invariant} \cite{du2020learning}.  Although  numerical results  in \cite{li2021invariant} \cite{du2020learning}  significantly outperform the state-of-the-art methods, the  methods  are heuristically motivated and lack theoretical justification.

In contrast to previous works, the key contributions of this paper are the following: 
\begin{itemize}
    \item We propose a new objective function that is motivated by the conditional entropy minimization (CEM) principle and show that it is explicitly related to the Deterministic Information Bottleneck (DIB) principle \cite{strouse2017deterministic}. 
    
    \item We theoretically show that under some  suitable  assumptions, minimizing the proposed objective function will  filter-out spurious features. 
    
    \item Our approach is general in the sense that it is  able to handle non-linear classifiers and may be extended to other DG methods that employ the invariance-principle. 
\end{itemize}

The key idea of our approach is to adopt the IRM framework for learning  a good representation function that can capture  both the true invariant features and the spurious invariant features, but penalize the conditional entropy of representations given labels  to filter-out the spurious invariant features.

The remainder of this paper is structured as follows. In Section \ref{sec: related work}, we summarize  relevant work  on DG and briefly introduce  the  IRM algorithm and  the  IB framework. In Section \ref{sec: problem setup} we formally define the problem and state and discuss the main assumptions underlying our theoretical analysis. Section \ref{sec: main results} provides the main theoretical results which motivate our practical approach proposed  in Section \ref{sec: practical method}. Experiments and their results are described in Section \ref{sec: numerical results} with concluding remarks presented in Section~\ref{sec: conclusion}.

\section{Related Work}
\label{sec: related work}
\subsection{Domain Generalization}
Numerous DG methods have been proposed in the past ten years which can be broadly categorized into some major directions, chiefly ``data manipulation'', representation learning, and meta-learning. The performance of a learning model often relies on the quantity and diversity of the training data and data manipulation is one of the cheapest methods to generate samples from a given set of limited data. Data manipulation can be employed via data augmentation \cite{nazari2020domain} \cite{borlino2021rethinking}, domain randomization \cite{khirodkar2019domain}, or adversarial data augmentation \cite{zhou2020deep}, \cite{yang2021adversarial}. The representation learning approach aims to learn a good representation feature by decomposing the prediction function into a representation function followed by a classifier. Over the past decade, many methods are emerged for better representation learning which can be categorized into two different learning principles: domain-invariant representation learning and feature disentanglement. Domain-invariant representation learning is based on the assumption that the representations are general and transferable to different domains if the representation features remain invariant from domain to domain \cite{ben2007analysis}. Notably, domain-invariant representation learning has emerged as one of the most common and efficient approaches in DG and provided many promising results \cite{arjovsky2019invariant} \cite{lu2021nonlinear} \cite{ahuja2021invariance} \cite{li2021invariant} \cite{du2020learning} \cite{ganin2016domain,mahajan2021domain,zhou2020domain,lyu2021barycentric,ahuja2020invariant}. Finally, meta-learning methods aim to learn the algorithm itself by learning from previous experience or tasks, i.e., learning-to-learn. Even though meta-learning is a general learning framework, it has recently been applied to DG tasks \cite{du2020learning} \cite{li2018learning} \cite{balaji2018metareg}. For more details, we refer the reader to the recent surveys on DG in \cite{wang2022generalizing} and \cite{zhou2021domain}. 

\subsection{Information Bottleneck and Invariant Risk Minimization}

In this section we review the IB framework \cite{tishby2000information}\cite{strouse2017deterministic} and the IRM algorithm \cite{arjovsky2019invariant} which are directly related to our proposed method. We use $f: \mathcal{X} \rightarrow \mathcal{Z}$ to denote a (potentially stochastic) representation mapping from the input data space $\mathcal{X}$ to the representation space $\mathcal{Z}$ and $g: \mathcal{Z} \rightarrow \mathcal{Y}$ to denote a classifier/labeling function from the representation space $\mathcal{Z}$ to the label space $\mathcal{Y}$.   
\subsubsection{Information Bottleneck Principle}
The IB method aims to find the best trade-off between accuracy and complexity (compression) when summarizing a random variable \cite{tishby2000information}. Particularly, IB aims to find a good (stochastic) representation function $f^*$ by solving the following optimization problem:
\begin{equation}
\label{eq: IBM loss}
    f^*= \argmin_{f} I(X;Z) - \theta I(Y;Z),
\end{equation} 
where 
$I(X;Z)$ denotes the mutual information between the  random variable  $X$ corresponding to input data and its representation $Z = f(X)$, $I(Y;Z)$ denotes the mutual information between the random variable $Y$ corresponding to the label and $Z$, and $\theta$ is a positive hyper-parameter that controls the trade off between maximizing $I(Y;Z)$ and minimizing $I(X;Z)$. 
Mutual information is a nonnegative statistical measure of dependence between random variables with larger values corresponding to stronger dependence and value zero corresponding to independence. Thus, the IB framework aims to find a representation $Z$ that is weakly dependent on input $X$, but is strongly dependent on the prediction label $Y$. Indirect rate-distortion source coding in information-theory provides an alternative interpretation of the IB objective with $Z$ viewed as a ``compressed'' encoding of $X$, $I(X;Z)$ is the number of ``bits'' needed to compress $X$ to $Z$, and $I(Y;Z)$ is a measure of how well the label $Y$ can be decoded from $Z$, i.e., a measure of prediction accuracy or ``inverse-distortion''. The IB problem can then be stated as a Lagrangian formulation of minimizing the number of bits needed to compress $X$ to $Z$ while being able to recover $Y$ from $Z$ to a desired accuracy.

The Deterministic Information Bottleneck (DIB) \cite{strouse2017deterministic} problem aims to find $f$ by solving the following optimization problem which is closely related to (\ref{eq: IBM loss}):
\begin{equation}
\label{eq: DIB loss}
    f^*= \argmin_{f} H(Z) - \theta I(Y;Z).
\end{equation}
For $\theta=1$, $H(Z) - \theta I(Y;Z)=H(Z|Y)$ which is the conditional entropy of the representation variable $Z$ given the label $Y$.  Thus, minimizing $H(Z|Y)$ is a special case of DIB where the aims of compression, i.e., minimizing $H(Z)$, and accuracy, i.e., maximizing $I(Y;Z)$, are equally weighted (balanced).

\subsubsection{Invariant Risk Minimization Algorithm}
The IRM algorithm \cite{arjovsky2019invariant} aims to find the representation $Z=f(X)$ for which the optimum classifier $g$ is invariant across all domains. The implicit assumption is that such representations and optimum domain-invariant classifiers exist. In practice, this is approximately realized by solving the following optimization problem \cite{arjovsky2019invariant}:
\begin{equation}
\label{eq: IRM loss}
       \min_{h \in \mathcal{G} \circ \mathcal{F}} \!\!L_{IRM}(h,\alpha) \!:=\! \sum_{i=1}^m \!\bigg[R^{(i)}(h) \!+\! \alpha \; \| \nabla_{t|t=1.0}R^{(i)}(t \cdot h) \|^2\bigg],
\end{equation}
where $\mathcal{F}$ is a family of representation functions (typically parameterized by weights of a neural network with a given architecture), $\mathcal{G}$ a family of \textit{linear} classifiers (typically the last fully connected classification layer of a classification neural network), $R^{(i)}(g \circ f) := \mathbb{E}_{(X,Y) \sim D_i} [\ell(g(f(X)), Y)]$ denotes a classification risk (e.g., error or cross-entropy loss) of using a representation function $f$ followed by a classifier $g$ in domain $i$ when using loss function $\ell$, and $\alpha$ is a hyper-parameter associated with the squared Euclidean norm of the gradients (denoted by $\nabla$) of the risks in different domains. When restricted to the family of linear classifiers and convex differentiable risk functions, Theorem 4 of \cite{arjovsky2019invariant} shows (under certain technical assumptions) that minimizing $L_{IRM}$ will yield a predictor that not only (approximately) minimizes the cumulative risk across all domains (the first term in $L_{IRM}$), but is also approximately optimum simultaneously across all domains, i.e., approximately invariant, and this is captured by the sum of squared risk gradients across all domains. 

In this paper, we rely on the IRM algorithm \cite{arjovsky2019invariant} to extract the invariant features and use the CEM principle to filter out the spurious invariant features. 
We note, however, that our approach is applicable to any method that can learn invariant features. We chose IRM due to its popularity and good empirical performance.  

\section{Problem Formulation}
\label{sec: problem setup}
 In this section, we formulate the minimum conditional entropy principle, which is a special case of the DIB principle, and show that it can be used to filter out spurious features. To do this we first introduce three modeling assumptions underlying our proposed approach. Our assumptions embrace two key ideas (i) the learned features are a linear mixture (superposition) of ``true'' domain-invariant features and ``spurious'' domain-specific features, and (ii) the invariant features are conditionally independent of spurious features given the label. 

\subsection{Notation}
Consider a classification task where the learning algorithm has access to i.i.d. data from the set of $m$ domains $\mathbb{D}=\{ D_1,D_2,\dots,D_m \}$.  The DG task is to learn a representation function $f: \mathcal{X} \rightarrow \mathcal{Z}$ from the input data space $\mathcal{X}$ to the representation space $\mathcal{Z}$, and a classifier $g: \mathcal{Z} \rightarrow \mathcal{Y}$ from the representation space $\mathcal{Z}$ to the label space $\mathcal{Y}$ that generalizes well to an unseen domain $D_s \notin \mathbb{D}$. 

Let $X$ denote the data random variable in input space, $Y$ the label random variable in label space, and $Z$ the extracted feature random variable in representation space.  Let the invariant and spurious features be denoted by $Z_{\textup{inv}}$ and $Z_{\textup{sp}}$, respectively. 
We denote expectation, variance, discrete/differential entropy, and mutual information by
$\mathbb{E}[\cdot]$, $\Var(\cdot)$, $H(\cdot)$, and $I(\cdot)$, respectively.

\subsection{Assumptions}

Ideally, we want to learn a representation function $f$ such that $f(X)=Z_{\textup{inv}}$. However, due to a finite number of observed domains, it is possible that the learned features might contain spurious invariant features which are invariant for all observed domains, but change in the unseen domain \cite{chen2021iterative} \cite{rosenfeld2021risks}. We model this situation by assuming that the representation function extracts features that are (approximately) composed of two elements: the (true) invariant features and the spurious invariant features: 
\begin{equation}
    f(X)= Z = \Theta(Z_{\textup{inv}},Z_{\textup{sp}}). \nonumber
\end{equation}
Next, we state three assumptions on $Z_{\textup{inv}},Z_{\textup{sp}}$ and $\Theta$ that we will use in Section~\ref{sec: main results} to derive our theoretical results. 

\begin{assumption}
\label{assumption: 1}
The (true) invariant features $Z_{\textup{inv}}$ are independent of the spurious invariant features $Z_{\textup{sp}}$ for a given label $Y$. Formally, $Z_{\textup{inv}} \indep Z_{\textup{sp}} | Y$. 
\end{assumption}

Assumption~\ref{assumption: 1} is widely accepted in the DG literature  \cite{nagarajan2020understanding} \cite{rosenfeld2021risks} \cite{ahuja2021invariance} \cite{neto2020causality}. For example, in the construction of the binary-MNIST dataset \cite{nagarajan2020understanding}, the class (label) is first selected, then the color (spurious feature) is independently added to the hand-written digit (invariant feature) picked from the selected class, making $Z_{\textup{inv}} \indep Z_{\textup{sp}} | Y$. For more details, we refer the reader to the third constraint in Section~3, page~5 of \cite{nagarajan2020understanding}.  In \cite{rosenfeld2021risks} \cite{ahuja2021invariance}  and \cite{neto2020causality}, this assumption is used but not explicitly stated. It is, however, implicit in Fig.~2 in \cite{ahuja2021invariance}, Fig.~3.1 in \cite{rosenfeld2021risks}, and the discussion below Fig.~2 in \cite{neto2020causality}. 

\begin{assumption}
\label{assumption: 2}
 The uncertainty of the invariant features is lower than the uncertainty of the spurious features when the label is known.  Formally, we assume $H(Z_{\textup{inv}}|Y) < H(Z_{\textup{sp}}|Y)$. 
\end{assumption}

Assumption~\ref{assumption: 2} has the following interesting clustering interpretation: invariant features are better clustered together in each class (have smaller variability) than spurious features. If additionally, $H(Z_{\textup{inv}}) = H(Z_{\textup{sp}})$, then $I(Z_{\textup{inv}};Y) = H(Z_{\textup{inv}}) - H(Z_{\textup{inv}}|Y) > H(Z_{\textup{sp}}) - H(Z_{\textup{sp}}|Y) = I(Z_{\textup{sp}};Y)$, implying that the invariant features $Z_{\textup{inv}}$ are more strongly related to the label $Y$ than the spurious features $Z_{\textup{sp}}$. 
 
\begin{assumption}
\label{assumption: 3}
$f(X)=Z=\Theta(Z_{\textup{inv}},Z_{\textup{sp}})=aZ_{\textup{inv}} + bZ_{\textup{sp}}$ and $\Var(Z|Y)=\Var(Z_{\textup{inv}}|Y)=\Var(Z_{\textup{sp}}|Y)=1$.
\end{assumption} 
Assumption \ref{assumption: 3} states that the extracted (learned) features are a linear combination of invariant features and spurious features, i.e.,  $Z=\Theta(Z_{\textup{inv}},Z_{\textup{sp}})=aZ_{\textup{inv}} + bZ_{\textup{sp}}$. This is similar in spirit to the settings in \cite{arjovsky2019invariant} \cite{ahuja2021invariance}
and is inspired by methods for Blind Source Separation (BSS) such as Independent Component Analysis (ICA) \cite{oja2004independent,hyvarinen2000independent,naik2011overview} which aim to separate-out statistically independent latent component sources, say $S_1, S_2, S_1 \indep S_2$, from observations of their \textit{linear combination} $M = a_1 S_1 + a_2 S_2$. Our focus on a simple linear combination model enables us to derive some insightful theoretical results in the next section and translate them into a practical algorithm for filtering-out spurious features in the context of domain generalization which provides substantial performance improvements over competing alternatives. A general non-linear dependence relationship $Z = \Theta(Z_{\textup{inv}},Z_{\textup{sp}})$ could potentially be handled using techniques such as non-linear ICA \cite{hyvarinen1999nonlinear} or non-linear IRM \cite{lu2021nonlinear} to filter out the spurious features. But we leave this to future work.

The assumption $\Var(Z|Y)=\Var(Z_{\textup{inv}}|Y)=\Var(Z_{\textup{sp}}|Y)=1$ is also motivated by an identical constraint in ICA needed to overcome the so-called \textit{scaling} ambiguity: if $S_1 \indep S_2$ and $M = a_1 S_1 + a_2 S_2$, then both $(S_1,S_2)$ and $(a_1S_1, a_2S_2)$ are pairs of independent component sources whose linear combination is $M$.
Finally, it is worth noting that Assumption~\ref{assumption: 1} and Assumption~\ref{assumption: 3} together imply that $a^2+b^2=1$ (see proof of Lemma \ref{lemma: 1}). 

\section{Main Results}
\label{sec: main results}
Our proposed approach is based on two fundamental steps. The first step is to extract all the invariant features $Z$ from source domains. These extracted invariant features may include both the true invariant features $Z_{\textup{inv}}$ and the spurious invariant features $Z_{\textup{sp}}$. The next step is to remove the spurious features in order to construct a classifier that purely relies on the true invariant features $Z_{\textup{inv}}$. For example, in the ``cow-camel setting'', the first step is to learn all extracted invariant features which might contain the color of the background. However, this spurious feature needs to be removed in the second step. 
In this section, we show that the CEM principle, i.e., minimizing $H(Z|Y)$, supports filtering-out the spurious invariant features.

\begin{assumption}
\label{assumption: 4}
Let 
\begin{eqnarray*}
       f^* &=& \argmin_{f} L_{\textup{invariant}}(f) \\
       s.t. \text{ \hspace{2 pt}   } H(f(X)|Y) &\leq& \gamma.
\end{eqnarray*}
where $L_{\textup{invariant}}$ is the loss function of an invariant representation learning algorithm. We assume that $L_{\textup{invariant}}$ is such that for all $\gamma$, $Z = f^*(X)$ is a linear superposition of both the invariant feature $Z_{\textup{inv}}$ and the spurious feature $Z_{\textup{sp}}$. 
\end{assumption}

Under Assumption~\ref{assumption: 4}, our key idea to ``eliminate'' the contribution of $Z_{\textup{sp}}$ from $Z$ by minimizing $L_{\textup{invariant}}$ subject to a suitable bound on the uncertainty of $Z$ given $Y$, i.e., designing a suitable value of $\gamma$. Indeed, we will show that there exists a suitable choice for $\gamma$ for which $f^*$ will extract only the (true) invariant feature $Z_{\textup{inv}}$ and filter-out $Z_{\textup{sp}}$. The key result needed to show this is the following lemma.

\begin{lemma}
\label{lemma: 1}
If Assumptions \ref{assumption: 1}, \ref{assumption: 2}, \ref{assumption: 3} hold, then
\begin{equation}
\label{eq: lemma_1}
    H(Z|Y)= H(a Z_{\textup{inv}} + b Z_{\textup{sp}}|Y) \geq H(Z_{\textup{inv}}|Y)
\end{equation}
and equality holds in (\ref{eq: lemma_1}) if, and only if, $a=1$ and $b=0$. 
\end{lemma}

\begin{proof}
Our proof of Lemma~\ref{lemma: 1} is for differential entropy, but it can be easily extended to discrete entropy (recall that we use $H(\cdot)$ to denote discrete or differential entropy). 
Under Assumptions \ref{assumption: 1} and \ref{assumption: 3}, we first show that $a^2+b^2=1$. Indeed, 
\begin{eqnarray}
    1 &=& \Var(Z|Y) = \Var(a Z_{\textup{inv}} + b Z_{\textup{sp}}|Y) \nonumber \\
      &=& a^2 \Var(Z_{\textup{inv}}|Y) + b^2 \Var(Z_{\textup{sp}}|Y) \label{eq: lemma 16}\\
      &=& a^2 +b^2, \label{eq: lemma 17}
\end{eqnarray}
where (\ref{eq: lemma 16}) is because $Z_{\textup{inv}} \indep Z_{\textup{sp}}|Y$ and (\ref{eq: lemma 17}) is due to the assumption that $\Var(Z_{\textup{inv}}|Y)=\Var(Z_{\textup{sp}}|Y)=1$. 
 
Next, we utilize the result in Lemma 1 of \cite{verdu2006simple} which states that for any two random variables $R_1$, $R_2$, and any two scalars $a$, $b$, if $R_1 \indep R_2$ and $a^2+b^2=1$, then:
\begin{eqnarray}
\label{eq: verdu}
    H(a R_1 + b R_2) \geq a^2 H(R_1) +b^2 H(R_2).
\end{eqnarray}

Now, for a given $Y=y \in \mathcal{Y}$, we have: 
\begin{eqnarray}
\lefteqn{H(aZ_{\textup{inv}}+bZ_{\textup{sp}}|Y=y)} \nonumber \\
    &\geq& a^2 H(Z_{\textup{inv}}|Y=y) + b^2 H(Z_{\textup{sp}}|Y=y)   \label{eq: lemma 11} \\
    &=& a^2 H(Z_{\textup{inv}}|Y=y) + b^2 H(Z_{\textup{inv}}|Y=y) \nonumber \\
    &+& b^2 H(Z_{\textup{sp}}|Y=y) - b^2 H(Z_{\textup{inv}}|Y=y) \nonumber \\
    &=& \!H(Z_{\textup{inv}}|Y\!=\!y) \nonumber \\
    &+& b^2 \! \big(H(Z_{\textup{sp}}|Y\!=\!y) \!-\! H(Z_{\textup{inv}}|Y\!=\!y)\! \big), \label{eq: lemma 12}
\end{eqnarray}
where (\ref{eq: lemma 11}) is due to (\ref{eq: verdu}) and $a^2+b^2=1$ and (\ref{eq: lemma 12}) is because $a^2+b^2=1$. Next,
\begin{eqnarray}
\lefteqn{H(Z|Y) = H(aZ_{\textup{inv}}+bZ_{\textup{sp}}|Y)}  \nonumber \\
    &=& \int_{y \in \mathcal{Y}} p(y)H(aZ_{\textup{inv}} + bZ_{\textup{sp}}|Y=y) \, dy \nonumber \\ 
    &\geq&  \int_{y \in \mathcal{Y}} p(y) H(Z_{\textup{inv}}|Y=y) \, dy \label{eq: lemma 13}\\
    &+& \int_{y \in \mathcal{Y}}p(y) b^2  \big(H(Z_{\textup{sp}}|Y\!=\!y) \!-\! H(Z_{\textup{inv}}|Y\!=\!y) \big) \, dy \nonumber \\ 
    &=& H(Z_{\textup{inv}}|Y) + b^2 \big(H(Z_{\textup{sp}}|Y)-H(Z_{\textup{inv}}|Y) \big) \label{eq: lemma 13-c} \\
    &\geq& H(Z_{\textup{inv}}|Y) \label{eq: lemma 14}
\end{eqnarray}
where (\ref{eq: lemma 13}) follows from (\ref{eq: lemma 12}) 
and (\ref{eq: lemma 14}) from $H(Z_{\textup{sp}}|Y) > H(Z_{\textup{inv}}|Y)$ (Assumption~\ref{assumption: 2}). If $a=1$ and $b=0$ then $Z = Z_{\textup{inv}}$ and equality holds. Conversely, if equality holds then $a=1$ and $b=0$ must hold, because otherwise we would have $b^2 > 0$ which together with $H(Z_{\textup{sp}}|Y) > H(Z_{\textup{inv}}|Y)$ and (\ref{eq: lemma 13-c}) would imply that $H(Z|Y)$ is strictly larger than $H(Z_{\textup{inv}}|Y)$. Thus,  equality $H(Z|Y)=H(Z_{\textup{inv}}|Y)$ occurs if, and only if, $a=1$ and $b=0$, or equivalently, if, and only if $Z=Z_{\textup{inv}}$. 
\end{proof}

Lemma \ref{lemma: 1} shows that $H(Z|Y)$ is always lower bounded by $H(Z_{\textup{inv}}|Y)$ and equality occurs if, and only if, $Z=Z_{\textup{inv}}$. We use Lemma~\ref{lemma: 1} to prove Theorem~\ref{theorem: 1} which states that the CEM principle can be used to extract the (true) invariant features $Z_{\textup{inv}}$.

\begin{theorem}
\label{theorem: 1}
If Assumptions \ref{assumption: 1}, \ref{assumption: 2}, \ref{assumption: 3}, and \ref{assumption: 4} hold, then there exits a $\gamma^*$ such that $f^*(X) = Z_{\textup{inv}}$.
\end{theorem}

\begin{proof}
From Assumption \ref{assumption: 4}, for any $\gamma$, minimizing $L_{\textup{invariant}}$ yields $Z = a Z_{\textup{inv}} + b Z_{\textup{sp}}$ for some values of $a,b$ that depend on $\gamma$. We also have $\gamma \geq H(Z|Y) \geq H(Z_{\textup{inv}}|Y)$, where the first inequality is due to the constraint in the minimization of $L_{\textup{invariant}}$ and the second is from Lemma~\ref{lemma: 1}. 
If we choose $\gamma = \gamma^* := H(Z_{\textup{inv}}|Y)$, then $H(Z|Y) = H(Z_{\textup{inv}}|Y)$. From Lemma \ref{lemma: 1}, $H(Z|Y) = H(Z_{\textup{inv}}|Y)$ if, and only if, $b=0$. Thus, selecting $\gamma^* = H(Z_{\textup{inv}}|Y)$ will lead to a representation function $f^*$ such that $f^*(X)=Z=Z_{\textup{inv}}$. 
\end{proof}

\section{Practical Approach}
\label{sec: practical method}

We propose to find invariant features by solving the following CEM optimization problem:
%
\begin{equation}
\label{eq: IBM and IRM objective function}
\min_{h \in \mathcal{G} \circ \mathcal{F}} L_{CE-IRM}(h,\alpha,\beta):= L_{IRM}(h,\alpha) + \beta H(f(X)|Y).
\end{equation}
This can be interpreted as the Lagrangian form of the optimization problem in Assumption~\ref{assumption: 4} with $L_{\textup{invariant}}$ replaced by the IRM loss function $L_{IRM}$ in (\ref{eq: IRM loss}) and the conditional entropy constraint in Assumption~\ref{assumption: 4} appearing as the second term with Lagrange multiplier $\beta$.
The two hyper-parameters  $\alpha$ and $\beta$ control the trade-off between minimizing the Invariant Risk loss and minimizing the conditional entropy loss.
Here, $Y$ denotes the label, $h = g \circ f$ acts as an invariant predictor 
with $f \in \mathcal{F}$, $g\in\mathcal{G}$, and $Z = f(X)$ is the output of the penultimate layer of the end-to-end neural network that implements $h = g\circ f$, i.e., the layer just before the output layer. We note that $Z$ and $Y$ represent, respectively, the latent representations and the labels corresponding to the input data $X$ from all seen domains \textit{combined}.

In order to practically solve the optimization problem in (\ref{eq: IBM and IRM objective function}), we leverage the implementations in \cite{ahuja2021invariance} and \cite{alemi2016deep}. 
Since
\begin{eqnarray*}
  H(Z|Y) =  H(Z) + H(Y|Z) -  H(Y)
\end{eqnarray*}
and $H(Y)$ is a data-dependent constant independent of $h = g \circ f$, the CEM optimization problem in (\ref{eq: IBM and IRM objective function}) is equivalent to the following one
\begin{equation}
\label{eq: CEM_alternative}
\min_{h \in \mathcal{G} \circ \mathcal{F}} L_{IRM}(h,\alpha) + \beta H(f(X)) + \beta H(Y|f(X)).
\end{equation}
The first two terms of the objective function in (\ref{eq: CEM_alternative}) are identical to the objective function proposed in \cite{ahuja2021invariance}. We therefore adapt the implementation in \cite{ahuja2021invariance}, which can be found at \href{https://github.com/ahujak/IB-IRM}{\underline{this link}} \footnote{\url{https://github.com/ahujak/IB-IRM}}, to minimize the first two terms in (\ref{eq: CEM_alternative}). 
In order to optimize the third conditional entropy term $H(Y|Z)$, we adopt the variational characterization of conditional entropy described in \cite{alemi2016deep}. A simple implementation of the variational method in \cite{alemi2016deep} for minimizing of conditional entropy is available at \href{https://github.com/1Konny/VIB-pytorch}{\underline{this link}}\footnote{\url{https://github.com/1Konny/VIB-pytorch}}.

\section{Experiments}\label{sec: numerical results}
\begin{table*}[ht]
\centering
\renewcommand{\arraystretch}{1.5}
\caption{Average accuracy  in percentage (\%)  of compared methods. The number of classes in LNU-3/3S and AC-CMNIST datasets is 2 while the number of classes in CS-CMNIST dataset is 10. ``\#Domains" denotes the number of domains in the dataset. 
\label{table: 1}}
\scalebox{1.0}{\resizebox{\columnwidth}{!}{
\begin{tabular}{c c c c c c c}
\hline
\textbf{Datasets} & \textbf{\#Domains}   & \textbf{ERM} \cite{vapnik1999overview} & \textbf{IRM} \cite{arjovsky2019invariant} & \textbf{IB-ERM} \cite{ahuja2021invariance} & \textbf{IB-IRM} \cite{ahuja2021invariance} & \textbf{CE-IRM (proposed)} \\ 
\hline
\textbf{CS-CMNIST} & 3 &  60.3 ± 1.2 & 61.5 ± 1.5   & 71.8 ± 0.7 &  71.8 ± 0.7 & \textbf{85.7 ± 0.9}    \\ 
\textbf{LNU-3} & 6 & 67.0 ± 18.0  & \textbf{86.0 ± 18.0}  &  74.0 ± 20.0 & 81.0 ± 19.0 & 84.0 ± 19.0    \\
\textbf{LNU-3S} & 6 & 64.0 ± 19.0  & 86.0 ± 18.0   & 73.0 ± 20.0 & 81.0 ± 19.0 & \textbf{90.0 ± 17.0}   \\
\textbf{LNU-3} & 3 & \textbf{52.0 ± 7.0}  & \textbf{52.0 ± 7.0}   & 51.0 ± 6.0 & \textbf{52.0 ± 7.0} &   \textbf{52.0 ± 7.0}  \\
\textbf{LNU-3S} & 3 &  51.0 ± 6.0  &  51.0 ± 7.0  &  51.0 ± 6.0 & 51.0 ± 7.0 & \textbf{52.0 ± 7.0}    \\
\textbf{AC-CMNIST} & 3 & 17.2 ± 0.6 & 16.5 ± 2.5 & 17.7 ± 0.5 &  \textbf{18.4 ± 1.4} &    17.5 ± 1.3     \\ 
\hline
\end{tabular}}}
\end{table*}

In this section, we evaluate the efficacy of our proposed method on some DG datasets that contain spurious features.

\subsection{Datasets}

\textbf{AC-CMNIST \cite{arjovsky2019invariant}}.  The Anti-causal-CMNIST dataset is a synthetic binary classification dataset derived from the MNIST dataset. It was proposed in \cite{arjovsky2019invariant} and is also used in \cite{ahuja2021invariance}.  
There are three domains in AC-CMNIST: two training domains containing 25,000 data points each, and one test domain containing 10,000 data points. 
Similar to the CMNIST dataset \cite{lecun1998mnist}, the images in AC-CMNIST are colored red or green in such a way that the color correlates strongly with the binary label in the two seen (training) domains, but is weakly correlated with the label in the unseen test domain. The goal is to identify whether the colored digit is less than five or more than five (binary label). Thus, in this dataset color is designed to be a spurious invariant feature.
For a fair comparison, we utilize the same construction of AC-CMNIST dataset as in \cite{arjovsky2019invariant} \cite{ahuja2021invariance}. 

\textbf{CS-CMNIST \cite{ahuja2020empirical}.} The Covariate-Shift-CMNIST dataset is a synthetic classification dataset derived from CMNIST dataset. It was proposed in \cite{ahuja2020empirical} and used in \cite{ahuja2021invariance}. This dataset has three domains: two training domains containing 20,000 data points each and one test domain also containing 20,000 data points.
We follow the construction method of \cite{ahuja2021invariance} to set up a ten-class classification task, where the ten classes are the ten digits from 0 to 9, and each digit class is assigned a color that is strongly correlated with the label in the two seen training domains and is independent of the label in the unseen test domain.   
Details of the CS-CMNIST and the model for generating this dataset can be found in Section~7.2.1.A of \cite{ahuja2020empirical}. For a fair comparison, we utilize the same construction methodology of the CS-CMNIST dataset as in \cite{ahuja2021invariance}. 

 \textbf{Linear unit dataset (LNU-3/3S) \cite{aubin2021linear}.} The linear unit (LNU) dataset is a synthesic dataset that is constructed from a linear low-dimensional model for evaluating out-of-distribution generalization algorithms under the effect of spurious invariant features \cite{aubin2021linear}. There are six sub-datasets in the LNU dataset, each sub-dataset consists of three or six domains, and each domain contains 10,000 data points. Due to limited time and space, we selected two sub-datasets from the LNU dataset named LNU-3 and LNU-3S to perform the evaluation. From the numerical results in \cite{ahuja2021invariance}, we note that LNU-3 and LNU-3S are the most challenging sub-datasets in the LNU dataset. 

\subsection{Methods Compared}
We compare our proposed method, named Conditional Entropy and Invariant Risk Minimization (CE-IRM) against the following competing alternatives: (i) Empirical Risk Minimization (ERM) \cite{vapnik1999overview} as a simple baseline, (ii) the original Invariant Risk Minimization (IRM) algorithm in \cite{arjovsky2019invariant}, (iii) the Information Bottleneck Empirical Risk Minimization (IB-ERM) algorithm in \cite{ahuja2021invariance}, and (iv) the Information Bottleneck Invariant Risk Minimization (IB-IRM) algorithm in \cite{ahuja2021invariance}. 
We omit comparison with the algorithm proposed in \cite{li2021invariant} since their implementation was not available at the time our paper was submitted. Moreover, with the exception of the CS-CMNIST dataset where our method improves over theirs about $10\%$ points, they do not report results for the other datasets that we used.

\subsection{Implementation Details}

We use the training-domain validation set tuning procedure in \cite{ahuja2021invariance} for tuning all hyper-parameters. To construct the validation set, we split the seen data into a training set and a validation set in the ratio of 95\% to 5\% and select the model that maximizes classification accuracy on the validation set. 

For AC-CMNIST, we utilize the learning model in  \cite{ahuja2021invariance} which is based on a simple Multi-Layer Perceptron (MLP) with two fully connected layers each having
an output size 256 followed by an output layer of size two which aims to identify whether the digit is less than 5 or more than 5. The Adam optimizer is used for training with a learning rate of $10^{-4}$, batch size of $64$, and the number of epochs set to $500$. To find the best representation, we search for the best values of weights of the Invariant Risk term and the Conditional Entropy term, i.e., $\alpha,  \beta$, respectively, among the following choices: $ 0.1, 1, 10, 10^2, 10^3, 10^4$.

For CS-CMNIST, we follow the learning model in \cite{ahuja2021invariance} which is composed of three convolutional layers with feature map dimensions of 256, 128, and 64. Each convolutional layer is followed by a ReLU activation and batch normalization layer. The last layer is a linear layer that aims to classify the digit to 10 classes. We use the SGD optimizer for training with a batch size of 128, learning rate of $10^{-1}$ with decay over every 600 steps, and the total number of steps set to 2,000. Similarly to AC-CMNIST, we perform a search for the weights of Invariant Risk and Conditional Entropy terms with $\alpha, \beta \in \{ 0.1, 1, 10, 10^2, 10^3, 10^4 \}$.

For the LNU dataset, we follow the procedure described in \cite{ahuja2021invariance}. Particularly, 20 pairs of $\alpha$ in the range $[1-10^{-0.3},1-10^{-3}]$, $\beta$ in the range $[1-10^{0}, 1-10^{-2}]$, learning rate in the range $[10^{-4}, 10^{-2}]$, and weight of decay in the range $[10^{-6},10^{-2}]$ are randomly sampled and trained. The best model is selected based on the training-domain validation set tuning procedure. 

We repeat the whole experiment five times by selecting five random seeds, where for each random seed, the whole process of tuning hyper-parameters and selecting models is repeated. Finally, the average accuracy and standard deviation values are reported.
The source code of our proposed algorithm is available at \href{https://github.com/thuan2412/Conditional_entropy_minimization_for_Domain_generalization}{\underline{this link}}.\footnote{\url{https://github.com/thuan2412/Conditional_entropy_minimization_for_Domain_generalization}}

\subsection{Results and Discussion}

The results of all our computer experiments are shown in Table~\ref{table: 1}. 
The numerical results of ERM, IRM, IB-ERM, and IB-IRM reported in Table~\ref{table: 1} are taken from \cite{ahuja2021invariance}. 
On the CS-CMNIST dataset, the four competing algorithms we tested achieve a classification accuracy in the range $60\% - 72\%$. But, our proposed CE-IRM algorithm vastly improves over the best alternative by almost 14\% points. This can be explained by the way the CS-CMNIST is generated. Indeed, by construction, the colors (spurious features) are added independently into the digits (invariant features) for a given label. Therefore our assumption $Z_{\textup{sp}} \indep Z_{\textup{inv}} |Y$ holds for the CS-CMNIST dataset. 

For the LNU dataset, we followed the procedures in \cite{ahuja2021invariance} to compute the classification error (equivalently accuracy) of the tested algorithms. We report the average accuracy together with its standard deviation in Table~\ref{table: 1}. Similarly to \cite{ahuja2021invariance}, we compare all algorithms on the LNU-3 dataset and the LNU-3S dataset with the number of domains set to 6 or 3 (we used the same 3 domains as in \cite{ahuja2021invariance}). 
For six domains, our CE-IRM algorithm outperforms all four competing methods by more than $4\%$ points on the LNU-3S dataset, but is only second-best on the LNU-3 dataset about $2\%$ point below the IRM algorithm. For three domains, the performance of all methods is very similar on both the LNU-3 and LNU-3S datasets. 
The results for the LNU-3 and LNU-3S datasets show that having more domains during training can improve the test accuracy of all algorithms.

Compared to the CS-CMNIST and the LNU-3/3S datasets, our results indicate that the AC-CMNIST is, by far, the most challenging dataset where none of the methods work well. Indeed, by construction, the AC-CMNIST contains strong spurious correlations between data and label leading to the failure of all tested algorithms. These results are consistent with those reported in \cite{arjovsky2019invariant}, \cite{ahuja2021invariance}, and \cite{li2021invariant}.

\section{Conclusions}
\label{sec: conclusion}
We proposed a new DG approach based on the CEM principle for filtering-out spurious features. Our practical implementation combines the well-known IRM algorithm and the CEM principle to achieve competitive or better performance compared to the state-of-the-art DG methods. In addition, we showed that our objective function is closely related to the DIB method, and theoretically proved that under certain conditions, our method can truly extract the invariant features. We focused on the simple model where the features learned by an IRM algorithm are a linear combination of true and spurious invariant features. Our future work will focus on combining the non-linear IRM algorithm \cite{lu2021nonlinear} with a nonlinear Blind Source Separation method, e.g., non-linear ICA \cite{hyvarinen1999nonlinear}, to accommodate non-linear mixture models of invariant features and spurious features.

\section{Acknowledgment}
The authors would like to acknowledge funding provided by AFOSR grant
\# FA9550-18-1-0465.

\balance
\bibliography{example_paper_2}

\begin{thebibliography}{10}
\providecommand{\url}[1]{#1}
\csname url@samestyle\endcsname
\providecommand{\newblock}{\relax}
\providecommand{\bibinfo}[2]{#2}
\providecommand{\BIBentrySTDinterwordspacing}{\spaceskip=0pt\relax}
\providecommand{\BIBentryALTinterwordstretchfactor}{4}
\providecommand{\BIBentryALTinterwordspacing}{\spaceskip=\fontdimen2\font plus
\BIBentryALTinterwordstretchfactor\fontdimen3\font minus
  \fontdimen4\font\relax}
\providecommand{\BIBforeignlanguage}[2]{{%
\expandafter\ifx\csname l@#1\endcsname\relax
\typeout{** WARNING: IEEEtran.bst: No hyphenation pattern has been}%
\typeout{** loaded for the language `#1'. Using the pattern for}%
\typeout{** the default language instead.}%
\else
\language=\csname l@#1\endcsname
\fi
#2}}
\providecommand{\BIBdecl}{\relax}
\BIBdecl

\bibitem{chen2021iterative}
Y.~Chen, E.~Rosenfeld, M.~Sellke, T.~Ma, and A.~Risteski, ``Iterative feature
  matching: Toward provable domain generalization with logarithmic
  environments,'' \emph{arXiv preprint arXiv:2106.09913}, 2021.

\bibitem{rosenfeld2021risks}
E.~Rosenfeld, P.~Ravikumar, and A.~Risteski, ``The risks of invariant risk
  minimization,'' in \emph{International Conference on Learning
  Representations}, vol.~9, 2021.

\bibitem{wang2022generalizing}
J.~Wang, C.~Lan, C.~Liu, Y.~Ouyang, T.~Qin, W.~Lu, Y.~Chen, W.~Zeng, and P.~Yu,
  ``Generalizing to unseen domains: A survey on domain generalization,''
  \emph{IEEE Transactions on Knowledge and Data Engineering}, 2022.

\bibitem{zhou2021domain}
K.~Zhou, Z.~Liu, Y.~Qiao, T.~Xiang, and C.~C. Loy, ``Domain generalization: A
  survey,'' \emph{arXiv preprint arXiv:2103.02503}, 2021.

\bibitem{arjovsky2019invariant}
M.~Arjovsky, L.~Bottou, I.~Gulrajani, and D.~Lopez-Paz, ``Invariant risk
  minimization,'' \emph{arXiv preprint arXiv:1907.02893}, 2019.

\bibitem{lu2021nonlinear}
C.~Lu, Y.~Wu, J.~M. Hern{\'a}ndez-Lobato, and B.~Sch{\"o}lkopf, ``Nonlinear
  invariant risk minimization: A causal approach,'' \emph{arXiv preprint
  arXiv:2102.12353}, 2021.

\bibitem{aubin2021linear}
B.~Aubin, A.~S{\l}owik, M.~Arjovsky, L.~Bottou, and D.~Lopez-Paz, ``Linear
  unit-tests for invariance discovery,'' \emph{arXiv preprint
  arXiv:2102.10867}, 2021.

\bibitem{kamath2021does}
P.~Kamath, A.~Tangella, D.~Sutherland, and N.~Srebro, ``Does invariant risk
  minimization capture invariance?'' in \emph{International Conference on
  Artificial Intelligence and Statistics}.\hskip 1em plus 0.5em minus
  0.4em\relax PMLR, 2021, pp. 4069--4077.

\bibitem{gulrajani2020search}
I.~Gulrajani and D.~Lopez-Paz, ``In search of lost domain generalization,'' in
  \emph{International Conference on Learning Representations}, 2020.

\bibitem{nagarajan2020understanding}
V.~Nagarajan, A.~Andreassen, and B.~Neyshabur, ``Understanding the failure
  modes of out-of-distribution generalization,'' in \emph{International
  Conference on Learning Representations}, 2020.

\bibitem{bui2021exploiting}
M.-H. Bui, T.~Tran, A.~Tran, and D.~Phung, ``Exploiting domain-specific
  features to enhance domain generalization,'' \emph{Advances in Neural
  Information Processing Systems}, vol.~34, 2021.

\bibitem{ahuja2021invariance}
K.~Ahuja, E.~Caballero, D.~Zhang, J.-C. Gagnon-Audet, Y.~Bengio, I.~Mitliagkas,
  and I.~Rish, ``Invariance principle meets information bottleneck for
  out-of-distribution generalization,'' \emph{Advances in Neural Information
  Processing Systems}, vol.~34, pp. 3438--3450, 2021.

\bibitem{tishby2000information}
N.~Tishby, F.~C. Pereira, and W.~Bialek, ``The information bottleneck method,''
  \emph{arXiv preprint physics/0004057}, 2000.

\bibitem{li2021invariant}
B.~Li, Y.~Shen, Y.~Wang, W.~Zhu, C.~J. Reed, J.~Zhang, D.~Li, K.~Keutzer, and
  H.~Zhao, ``Invariant information bottleneck for domain generalization,''
  \emph{arXiv preprint arXiv:2106.06333}, 2021.

\bibitem{du2020learning}
Y.~Du, J.~Xu, H.~Xiong, Q.~Qiu, X.~Zhen, C.~G. Snoek, and L.~Shao, ``Learning
  to learn with variational information bottleneck for domain generalization,''
  in \emph{European Conference on Computer Vision}.\hskip 1em plus 0.5em minus
  0.4em\relax Springer, 2020, pp. 200--216.

\bibitem{strouse2017deterministic}
D.~Strouse and D.~J. Schwab, ``The deterministic information bottleneck,''
  \emph{Neural computation}, vol.~29, no.~6, pp. 1611--1630, 2017.

\bibitem{nazari2020domain}
N.~H. Nazari and A.~Kovashka, ``Domain generalization using shape
  representation,'' in \emph{European Conference on Computer Vision}.\hskip 1em
  plus 0.5em minus 0.4em\relax Springer, 2020, pp. 666--670.

\bibitem{borlino2021rethinking}
F.~C. Borlino, A.~D'Innocente, and T.~Tommasi, ``Rethinking domain
  generalization baselines,'' in \emph{2020 25th International Conference on
  Pattern Recognition (ICPR)}.\hskip 1em plus 0.5em minus 0.4em\relax IEEE,
  2021, pp. 9227--9233.

\bibitem{khirodkar2019domain}
R.~Khirodkar, D.~Yoo, and K.~Kitani, ``Domain randomization for scene-specific
  car detection and pose estimation,'' in \emph{2019 IEEE Winter Conference on
  Applications of Computer Vision (WACV)}.\hskip 1em plus 0.5em minus
  0.4em\relax IEEE, 2019, pp. 1932--1940.

\bibitem{zhou2020deep}
K.~Zhou, Y.~Yang, T.~Hospedales, and T.~Xiang, ``Deep domain-adversarial image
  generation for domain generalisation,'' in \emph{Proceedings of the AAAI
  Conference on Artificial Intelligence}, vol.~34, no.~07, 2020, pp.
  13\,025--13\,032.

\bibitem{yang2021adversarial}
F.-E. Yang, Y.-C. Cheng, Z.-Y. Shiau, and Y.-C.~F. Wang, ``Adversarial
  teacher-student representation learning for domain generalization,''
  \emph{Advances in Neural Information Processing Systems}, vol.~34, 2021.

\bibitem{ben2007analysis}
S.~Ben-David, J.~Blitzer, K.~Crammer, F.~Pereira \emph{et~al.}, ``Analysis of
  representations for domain adaptation,'' \emph{Advances in neural information
  processing systems}, vol.~19, p. 137, 2007.

\bibitem{ganin2016domain}
Y.~Ganin, E.~Ustinova, H.~Ajakan, P.~Germain, H.~Larochelle, F.~Laviolette,
  M.~Marchand, and V.~Lempitsky, ``Domain-adversarial training of neural
  networks,'' \emph{The journal of machine learning research}, vol.~17, no.~1,
  pp. 2096--2030, 2016.

\bibitem{mahajan2021domain}
D.~Mahajan, S.~Tople, and A.~Sharma, ``Domain generalization using causal
  matching,'' in \emph{International Conference on Machine Learning}.\hskip 1em
  plus 0.5em minus 0.4em\relax PMLR, 2021, pp. 7313--7324.

\bibitem{zhou2020domain}
F.~Zhou, Z.~Jiang, C.~Shui, B.~Wang, and B.~Chaib-draa, ``Domain generalization
  with optimal transport and metric learning,'' \emph{arXiv preprint
  arXiv:2007.10573}, 2020.

\bibitem{lyu2021barycentric}
B.~Lyu, T.~Nguyen, P.~Ishwar, M.~Scheutz, and S.~Aeron, ``Barycentric-alignment
  and invertibility for domain generalization,'' \emph{arXiv preprint
  arXiv:2109.01902}, 2021.

\bibitem{ahuja2020invariant}
K.~Ahuja, K.~Shanmugam, K.~Varshney, and A.~Dhurandhar, ``Invariant risk
  minimization games,'' in \emph{International Conference on Machine
  Learning}.\hskip 1em plus 0.5em minus 0.4em\relax PMLR, 2020, pp. 145--155.

\bibitem{li2018learning}
D.~Li, Y.~Yang, Y.-Z. Song, and T.~Hospedales, ``Learning to generalize:
  Meta-learning for domain generalization,'' in \emph{Proceedings of the AAAI
  Conference on Artificial Intelligence}, vol.~32, no.~1, 2018.

\bibitem{balaji2018metareg}
Y.~Balaji, S.~Sankaranarayanan, and R.~Chellappa, ``Metareg: Towards domain
  generalization using meta-regularization,'' \emph{Advances in Neural
  Information Processing Systems}, vol.~31, pp. 998--1008, 2018.

\bibitem{neto2020causality}
E.~C. Neto, ``Causality-aware counterfactual confounding adjustment for feature
  representations learned by deep models,'' \emph{arXiv preprint
  arXiv:2004.09466}, 2020.

\bibitem{oja2004independent}
E.~Oja and A.~Hyvarinen, ``Independent component analysis: A tutorial,''
  \emph{Helsinki University of Technology, Helsinki}, 2004.

\bibitem{hyvarinen2000independent}
A.~Hyv{\"a}rinen and E.~Oja, ``Independent component analysis: algorithms and
  applications,'' \emph{Neural networks}, vol.~13, no. 4-5, pp. 411--430, 2000.

\bibitem{naik2011overview}
G.~R. Naik and D.~K. Kumar, ``An overview of independent component analysis and
  its applications,'' \emph{Informatica}, vol.~35, no.~1, 2011.

\bibitem{hyvarinen1999nonlinear}
A.~Hyv{\"a}rinen and P.~Pajunen, ``Nonlinear independent component analysis:
  Existence and uniqueness results,'' \emph{Neural networks}, vol.~12, no.~3,
  pp. 429--439, 1999.

\bibitem{verdu2006simple}
S.~Verd{\'u} and D.~Guo, ``A simple proof of the entropy-power inequality,''
  \emph{IEEE Transactions on Information Theory}, vol.~52, no.~5, pp.
  2165--2166, 2006.

\bibitem{alemi2016deep}
A.~A. Alemi, I.~Fischer, J.~V. Dillon, and K.~Murphy, ``Deep variational
  information bottleneck,'' \emph{arXiv preprint arXiv:1612.00410}, 2016.

\bibitem{vapnik1999overview}
V.~N. Vapnik, ``An overview of statistical learning theory,'' \emph{IEEE
  transactions on neural networks}, vol.~10, no.~5, pp. 988--999, 1999.

\bibitem{lecun1998mnist}
Y.~LeCun, ``The mnist database of handwritten digits,'' \emph{http://yann.
  lecun. com/exdb/mnist/}, 1998.

\bibitem{ahuja2020empirical}
K.~Ahuja, J.~Wang, A.~Dhurandhar, K.~Shanmugam, and K.~R. Varshney, ``Empirical
  or invariant risk minimization? a sample complexity perspective,'' in
  \emph{International Conference on Learning Representations}, 2020.

\end{thebibliography}
\bibliographystyle{IEEEtran}
\end{document}